\documentclass[journal]{IEEEtran}
\usepackage{graphicx}
\usepackage{booktabs}
\usepackage{multirow}

\usepackage[accsupp]{axessibility} 
\usepackage{hyperref}
\usepackage{pifont}
\usepackage{mathtools}
\usepackage[numbers,square,comma,sort&compress]{natbib}
\usepackage{orcidlink}
\usepackage{algorithmic}
\usepackage{algorithm}
\ifCLASSINFOpdf
\else
\fi
\begin{document}
%
\title{WaterVideoQA: ASV-Centric Perception and Rule-Compliant Reasoning via Multi-Modal Agents}
%
%
%

\author{Runwei Guan,
        Shaofeng Liang,
        Ningwei Ouyang,
        Weichen Fei,
        Shanliang Yao,
        Wei Dai,
        Chenhao Ge,
        Penglei Sun,
        Xiaohui Zhu,
        Tao Huang,
        Ryan Wen Liu,
        and~Hui~Xiong,~\IEEEmembership{~Fellow,~IEEE}
\thanks{Runwei Guan, Shaofeng Liang, Penglei Sun and Hui Xiong are with Thrust of Artificial Intelligence, The Hong Kong University of Science and Technology (Guangzhou), Guangzhou, China.}
\thanks{Runwei Guan and Wen Liu are with Hubei Key Laboratory of Inland Shipping Technology (Wuhan University of Technology), Wuhan, China; Wen Liu is also with School of Navigation, Wuhan University of Technology, Wuhan, China.}
\thanks{Ningwei Ouyang, Wei Dai and Xiaohui Zhu are with School of Advanced Technology, Xi'an Jiaotong-Liverpool University, Suzhou, China.}
\thanks{Weichen Fei is with School of Artificial Intelligence, Nanjing University, Nanjing, China.}
\thanks{Shanliang Yao is with the School of Information Engineering, Yancheng Institute of Technology, Yancheng, China.}
\thanks{Chenhao Ge is with School of Engineering, Stanford University, Palo Alto, USA.}
\thanks{Tao Huang is with the Centre for AI and Data Science Innovation and the School of Science and Engineering, James Cook University, Smithfield, Australia.}
\thanks{Corresponding authors: xionghui@hkust-gz.edu.cn, wenliu@whut.edu.cn.}}

\maketitle

\begin{abstract}
While autonomous navigation has achieved remarkable success in passive perception (e.g., object detection and segmentation), it remains fundamentally constrained by a void in knowledge-driven, interactive environmental cognition. In the high-stakes domain of maritime navigation, the ability to bridge the gap between raw visual perception and complex cognitive reasoning is not merely an enhancement but a critical prerequisite for Autonomous Surface Vessels to execute safe and precise maneuvers. To this end, we present WaterVideoQA, the first large-scale, comprehensive Video Question Answering benchmark specifically engineered for all-waterway environments. This benchmark encompasses 3,029 video clips across six distinct waterway categories, integrating multifaceted variables such as volatile lighting and dynamic weather to rigorously stress-test ASV capabilities across a five-tier hierarchical cognitive framework. Furthermore, we introduce NaviMind, a pioneering multi-agent neuro-symbolic system designed for open-ended maritime reasoning. By synergizing Adaptive Semantic Routing, Situation-Aware Hierarchical Reasoning, and Autonomous Self-Reflective Verification, NaviMind transitions ASVs from superficial pattern matching to regulation-compliant, interpretable decision-making. Experimental results demonstrate that our framework significantly transcends existing baselines, establishing a new paradigm for intelligent, trustworthy interaction in dynamic maritime environments.
\end{abstract}


%
\IEEEpeerreviewmaketitle

\section{Introduction}
\label{sec:intro}

Driven by the rapid evolution of deep learning, Autonomous Surface Vessels (ASVs) and autonomous ship navigation have witnessed significant advancements \cite{han2026survey,yao2023radar}, particularly within the realm of intelligent perception \cite{trinh2025comprehensive}. Existing methodologies have demonstrated remarkable proficiency in foundational tasks such as object detection \cite{guan2023achelous,kristan2015fast,guan2026achelous++}, semantic segmentation \cite{guan2024mask,guan2024asy,sun2025caraffusion} and multi-object tracking \cite{usvtrack}. However, these capabilities are predominantly confined to a passive perception paradigm, where the system acts as a static observer of visual patterns. This reliance on superficial feature matching lacks the active cognition necessary to decode the underlying causalities and dynamic interactions of complex waterborne environments \cite{peng2019output}. In maritime scenarios, characterized by irregular waterway geometries, volatile tides, and unpredictable obstacle movements, relying solely on passive entity recognition is intrinsically insufficient \cite{jing2025energy}. Without a transition toward knowledge-driven reasoning and interactive environmental cognition, current ASVs remain incapable of interpreting high-level navigational rules or elucidating the causal rationale behind necessary maneuvers, thereby severely restricting their deployment in safety-critical autonomous operations \cite{liu2020scanning}.

\begin{figure*}
    \centering
    \includegraphics[width=0.998\linewidth]{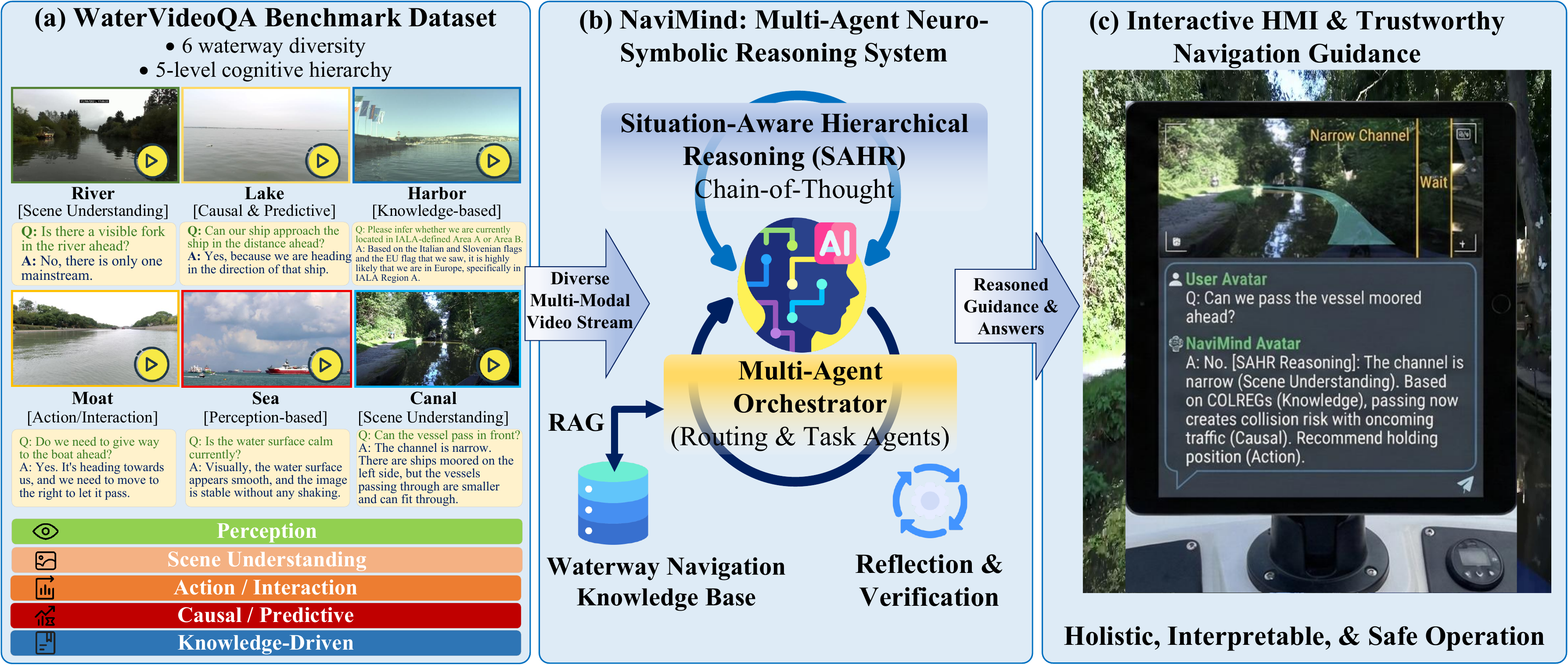}
    \vspace{-6mm}
    \caption{The overview of pipeline, including (a) the proposed WaterVideoQA dataset; (b) the proposed multi-agent neuro-symbolic reasoning system: NaviMind; (c) the real-word application scenarios for trustworthy navigation guidance.}
    \label{fig:pipeline}
\end{figure*}

Based on comprehensive previous research, the characteristics of open water environments can be summarized as follows: (1) Waterways, especially inland rivers, often feature irregular environments \cite{wang2025inland}. (2) The flow velocity in inland waters and oceanic tides cause some obstacles to be in a state of frequent movement \cite{li2025end}. (3) Waterway navigation lacks strictly fixed routes, accompanied by a relatively high frequency of unpredictable events \cite{rosic2025semantic}. (4) The significant contextual disparity between inland and marine environments causes considerably different model paradigms.
Consequently, although current ASVs can excel in areas like object detection through perception algorithms, they fundamentally lack a deep understanding of dynamic interactions, causal relationships, and complex navigation rules \cite{maza2022colregs}. For instance, merely knowing ``there is a vessel ahead" is insufficient for making collision avoidance decisions; the ASV must understand that ``the vessel is navigating through a narrow channel, and according to the rules, I must give way."


To advance semantic understanding, recent efforts have introduced natural language for interactive maritime perception, such as text-guided visual grounding in WaterVG \cite{guan2025watervg} and surveillance captioning in Da Yu \cite{guan2025yu}. Despite moving beyond pure coordinate-based detection, these paradigms are severely constrained by a single-frame bottleneck. Static snapshots inherently fail to capture the continuous motion trends crucial for causal reasoning and collision risk assessment. Furthermore, devoid of professional navigation rules, their comprehension relies on superficial visual pattern matching. Consequently, they are restricted to basic object identification and spatial localization, failing to execute the critical causal and procedural reasoning required for autonomous maneuvering and logical follow-up inquiries. Lastly, these studies are strictly confined to inland waters, neglecting the contextual complexities of open-sea scenarios.

To bridge the gap between passive perception and active cognition, we pioneer WaterVideoQA, the first comprehensive VideoQA benchmark for waterway environmental understanding. Overcoming the limitations of static or singular-scenario datasets, it integrates temporal dynamics across diverse inland and marine domains. We establish a rigorous five-tier cognitive hierarchy: Perception, Understanding, Behavior, Causality, and Knowledge-Driven Reasoning, to evaluate a model’s transition from basic object recognition to complex, rule-based navigational deduction. This provides a standardized metric for assessing ASV interpretability and decision-making depth in high-stakes maritime operations.

\begin{table*}
  \caption{A comprehensive comparison of waterway datasets. The name of \textbf{T-Levels} is Task-Levels, including perception (P), scene understanding (S), causal prediction (C), action or interaction (A) and reasoning (R). \textbf{EKR} denotes external knowledge retrieval. The upper and lower row in \textbf{Vocab Size} column denote the sizes of total and unique vocabulary.}
    \vspace{-3mm}
  \label{tab:datasets}
      \setlength\tabcolsep{6.8pt}
        \begin{tabular}{cc|ccccccc}
          \toprule
          \textbf{Datasets} & \textbf{Venues} & \textbf{Tasks} & \textbf{Temporal} & \textbf{Reasoning} & \textbf{EKR} & \textbf{Scenarios} & \textbf{Videos / Frames} & \textbf{Vocab Size}\\
          \midrule
          SMD \cite{moosbauer2019benchmark} & CVPRW$_\text{2019}$ & P & \ding{51} & \ding{55} & \ding{55} & Sea &  63 clips / 31.6k frames & N/A \\
          Flow \cite{cheng2021flow} & ICCV$_\text{2021}$ & P & \ding{51} & \ding{55} & \ding{55} & River & 200 clips / 54.6k frames & N/A \\
          LaRS-Seq \cite{vzust2023lars} & ICCV$_\text{2023}$ & P & \ding{51} & \ding{55} & \ding{55} & Sea, River, Lake & - / 53.3k frames & N/A \\
          \multirow{2}{*}{USVTrack} & \multirow{2}{*}{IROS$_{\text{2025}}$} & \multirow{2}{*}{P} & \multirow{2}{*}{\ding{51}} & \multirow{2}{*}{\ding{55}} & \multirow{2}{*}{\ding{55}} & River, Lake & \multirow{2}{*}{60 clips / 68.8k frames} & \multirow{2}{*}{N/A} \\
           & & & & & & Canal, Moat, Dock & \\
          \midrule
          \multirow{2}{*}{WaterVG \cite{guan2025watervg}} & \multirow{2}{*}{TITS$_{\text{2025}}$} & \multirow{2}{*}{P, S, R} & \multirow{2}{*}{\ding{55}} & \multirow{2}{*}{\ding{51}} & \multirow{2}{*}{\ding{55}} & River, Lake & \multirow{2}{*}{N/A} & 133.9K \\
           &  &  &  &  &  & Canal, Moat, Dock & & 1.80K\\
           \multirow{2}{*}{WaterCap \cite{guan2025yu}} & \multirow{2}{*}{TCSVT$_{\text{2026}}$} & \multirow{2}{*}{P, S} & \multirow{2}{*}{\ding{55}} & \multirow{2}{*}{\ding{55}} & \multirow{2}{*}{\ding{55}} & River, Lake & \multirow{2}{*}{N/A} & 1.88M \\
          &  &  &  &  &  & Canal, Moat, Dock & & 3.44K \\
          \midrule
           \multirow{2}{*}{WaterVideoQA} & \multirow{2}{*}{\textbf{2026}} & \textbf{P, S,} & \multirow{2}{*}{\ding{51}} & \multirow{2}{*}{\ding{51}} & \multirow{2}{*}{\ding{51}} & \textbf{River, Lake, Canal} & \multirow{2}{*}{\textbf{3k clips / 364.4k frames}} & \textbf{94.0K}\\
            & & \textbf{C, A, R} & & & & \textbf{Moat, Harbor, Sea} &  & \textbf{2.51K}\\
          \bottomrule
        \end{tabular}
\end{table*}

Overall, the contributions of the paper are summarized as:
\begin{enumerate}
    \item The first VideoQA dataset termed WaterVideoQA, oriented at all-waterway navigation, which contains 3,029 videos and 3,673 QA pairs with five hierarchical types.
    
    \item The first multi-agent system called NaviMind, which can efficiently and reliably understand the precious spatial and temporal information in the video and give the answer with high reliability.
    \item A novel Situation-Aware Hierarchical Reasoning mechanism equipped with adaptive retrieval-augmented generation, which enables the system to anchor visual evidence, dynamically retrieve professional maritime regulations, and align visual perceptions with rules.
    \item An efficient Autonomous Self-Reflective Verification mechanism for hallucination mitigation. NaviMind can self-diagnose potential hallucinations and trigger forced reasoning corrections, ensuring that outputs are not only logically coherent but also strictly visually grounded and rule-compliant.
\end{enumerate}

The remainder of this paper is organized as follows. Section \ref{sec:related} reviews the related work in waterway perception and multi-modal agents. Section \ref{sec:dataset} presents the construction, statistics, and characteristics of the proposed WaterVideoQA benchmark. Section \ref{sec:method} elaborates on the architectural details of the multi-agent NaviMind system. Section \ref{sec:experiments} provides a comprehensive experimental evaluation and analysis. Finally, Section \ref{sec:limit} discusses current limitations and future directions, followed by the conclusion in Section \ref{sec:conclusion}.

\section{Related Works}
\label{sec:related}

\subsection{Waterway Passive Perception based on Deep Learning}
Deep learning-based perception paradigms (e.g., detection, segmentation) are widely applied to ASVs \cite{trinh2025comprehensive,bovcon2021wasr}. For instance, Achelous++ \cite{guan2026achelous++} addresses multi-task perception, while Mask-VRDet \cite{guan2024mask} and ASY-VRNet \cite{guan2024asy} utilize vision-radar fusion. However, restricted to superficial pattern matching, these closed-set systems remain confined to passive perception \cite{wang2025inland}, lacking a deep understanding of dynamic interactions and complex navigation rules. To bridge this gap toward active cognition, we propose WaterVideoQA and the NaviMind multi-agent system. Via Situation-Aware Hierarchical Reasoning, NaviMind dynamically integrates real-time visual observations with retrieved maritime regulations , shifting the paradigm from mere obstacle identification to interpretable, regulation-compliant reasoning in open-ended scenarios.

\subsection{Waterway Understanding with Natural Language}
Natural language provides discrete symbolic signals that can supervise perceptual sensors (e.g., vision) and mediate human-machine interactive perception. While Vision-Language Models (VLMs) have advanced driving behavior explanation, human-vehicle communication, and end‑to‑end autonomous driving for ground \cite{cui2024survey,zhou2024vision} and aerial vehicles \cite{sun2026autofly}, progress in ASVs remains scarce. As Table \ref{tab:datasets} presents, recent efforts have started to explore language‑guided waterway understanding: WaterVG \cite{guan2025watervg} introduces the first visual‑grounding dataset for waterways by integrating multi‑sensor cues; NanoMVG \cite{guan2025nanomvg} deploys a lightweight grounding model on edge devices for ASVs; VLMAR \cite{wang2025vlmar} employs VLMs and MLLMs for maritime anomaly retrieval; and WaterCaption \cite{guan2025yu} constructs a large‑scale captioning dataset with a baseline MLLM. Nevertheless, these works exhibit two fundamental limitations: \textbf{(1) Single‑frame bottleneck:} decisions and trajectory predictions require temporal context, yet existing datasets rely on static images; \textbf{(2) Narrow scenario coverage:} current datasets are restricted to either inland or maritime settings, preventing models from learning domain‑agnostic representations and adaptable rule‑based reasoning across diverse visual dynamics, traffic patterns, and regulatory regimes. To overcome these gaps, we present WaterVideoQA, the first video‑based question‑answering benchmark spanning rivers, lakes, canals, moats, harbors, and seas, with comprehensive Q\&A categories that reflect real‑world navigation needs.

\begin{figure*}
  \begin{center}
    \centerline{\includegraphics[width=0.98\linewidth]{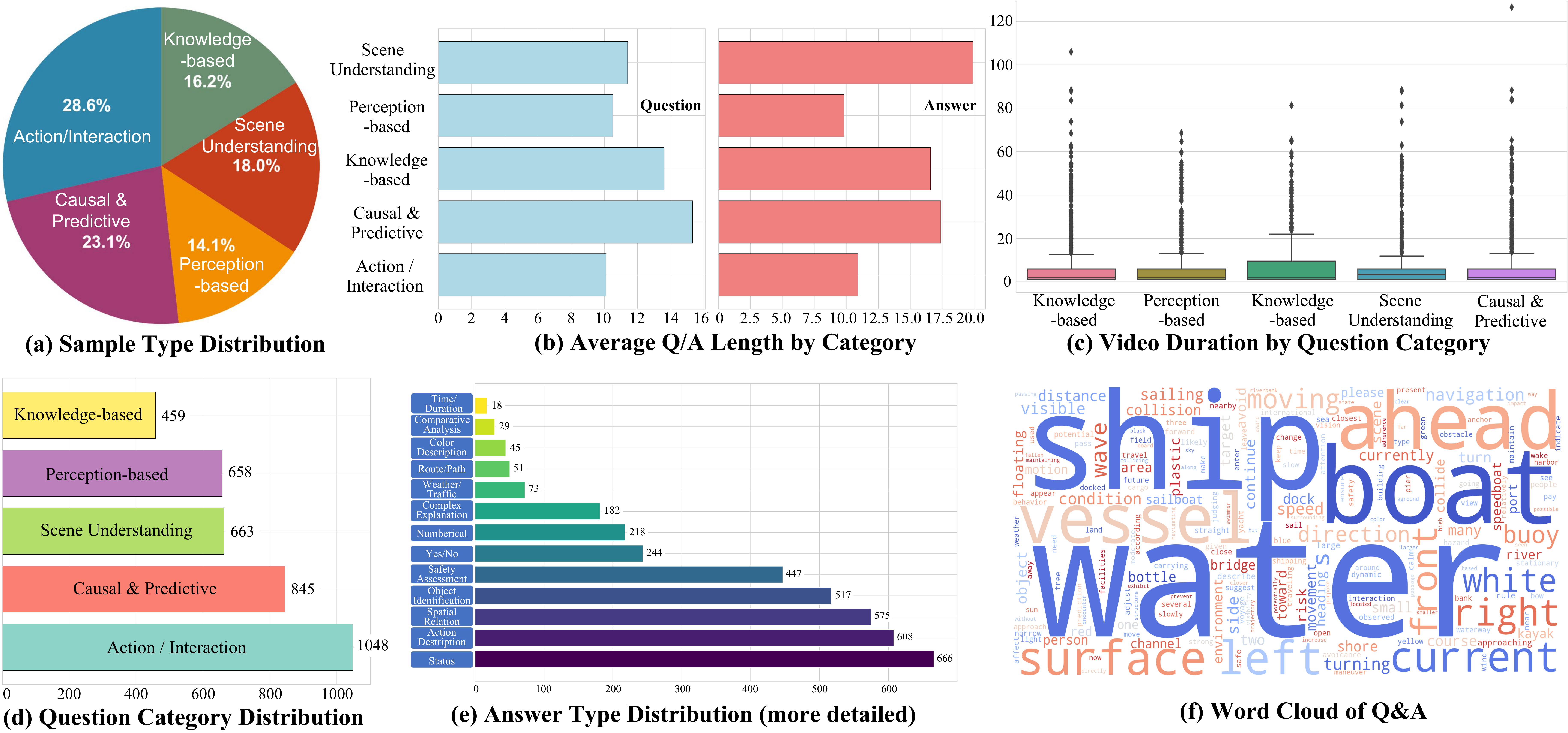}}
    \vspace{-3mm}
    \caption{The statistics of our proposed WaterVideoQA dataset, including (a) Sample Type Distribution, (b) Average Q/A Length by Category, (c) Video Duration by Question Category, (d) Question Category Distribution, (e) Answer Type Distribution and (f) Word Cloud of Q\&A.}
    \label{fig:dataset_stat}
  \end{center}
\end{figure*}

\begin{figure}
    \centering
    \includegraphics[width=0.99\linewidth]{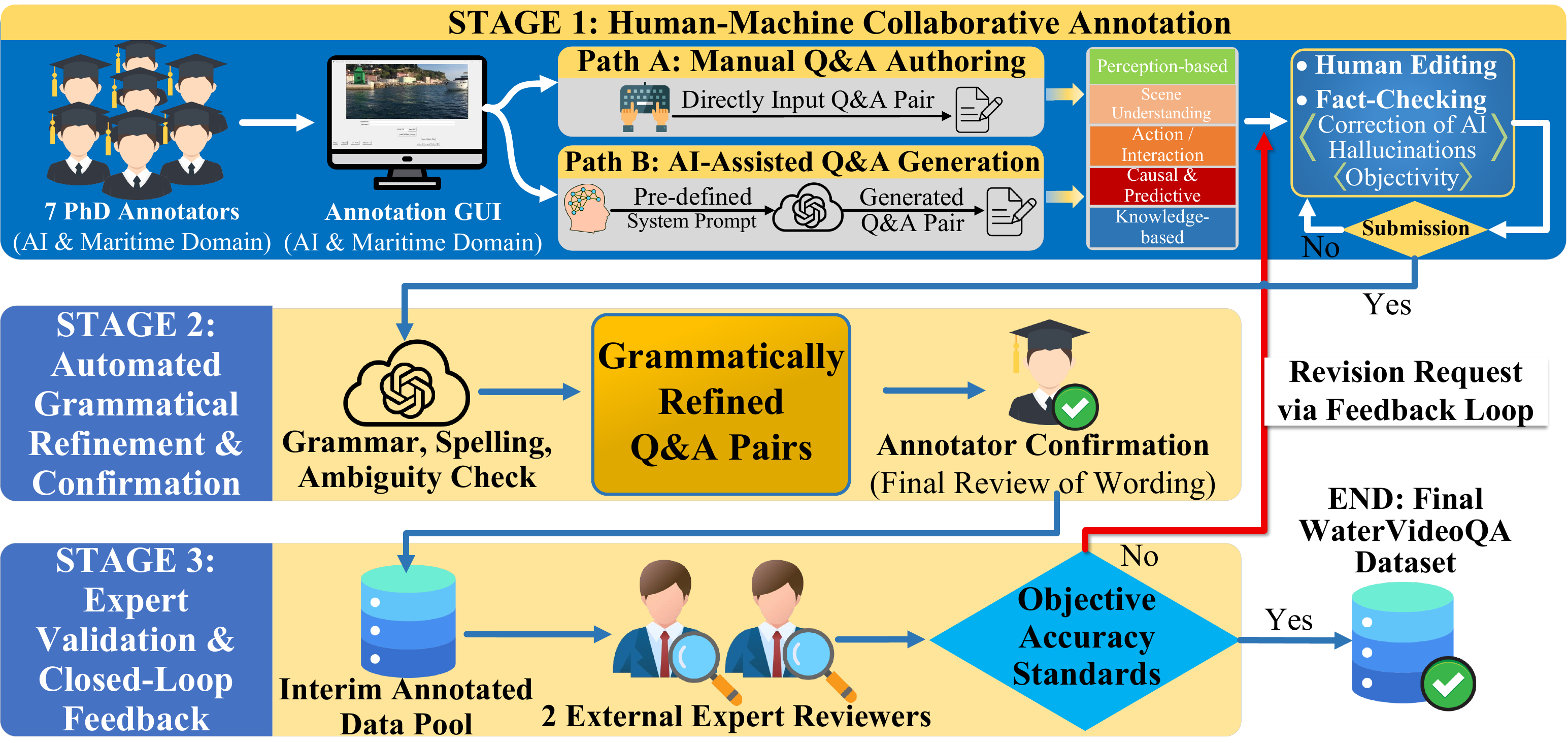}
    \vspace{-6mm}
    \caption{The overview of annotation process for WaterVideoQA.}
    \label{fig:annotation_process}
\end{figure}

\subsection{Video Understanding upon Multi-Agent System}
Video understanding has evolved from traditional spatio-temporal feature aggregation \cite{selva2023video,liang2025cognitive} to Large Language Model (LLM)-driven paradigms \cite{fan2024videoagent}. State-of-the-art MLLMs (e.g., Video-LLaVA \cite{lin2024video}, VideoChat \cite{li2025videochat}) excel in open-ended VideoQA by aligning visual encoders with LLM reasoning. However, these general-purpose models often suffer from hallucinations and lack the domain-specific logic critical for safety-critical autonomous navigation.
To handle complex reasoning, Multi-Agent Systems (MAS) assign distinct roles to simulate human cognitive processes via collaboration \cite{wu2025multi}. Recent works apply MAS to embodied planning and driving scene understanding \cite{xu2024drivegpt4}, decomposing intricate visual inputs into executable actions. Despite these advancements, applying MAS to maritime environments remains unexplored. Unlike structured urban roads, highly dynamic waterways require strict adherence to professional maritime regulations. Existing video reasoning frameworks \cite{fan2024videoagent,lin2024video} struggle to logically align visual observations with these domain-specific textual rules, yielding superficial explanations that fail to elucidate the causal rationale behind decisions. To bridge this gap, we introduce NaviMind, a multi-agent framework integrating a Situation-Aware Hierarchical Reasoning (SAHR) mechanism. By synergizing visual perception and knowledge-retrieval agents, NaviMind executes rigorous, rule-compliant semantic reasoning across all-waterway environments.

\section{WaterVideoQA Dataset}
\label{sec:dataset}

To bridge the gap between passive perception and active reasoning in maritime environments, we present WaterVideoQA, the first VideoQA benchmark for holistic waterway understanding. Curated from renowned temporal datasets: USVTrack \cite{usvtrack}, LaRS \cite{vzust2023lars}, Flow \cite{cheng2021flow}, MODD2 \cite{bovcon2018stereo}, and SMD \cite{moosbauer2019benchmark}, our benchmark comprehensively covers real-world complexities. This section details our human-machine collaborative annotation pipeline, designed to guarantee objective accuracy, followed by an analysis of dataset statistics and diversity.

\subsection{Annotation Process}
To construct a high-quality benchmark that faithfully reflects the complexities of real-world waterway environments, we orchestrated a rigorous Human-Machine Collaborative Annotation pipeline. This process was designed to mitigate the hallucinations common in general-purpose LLMs while leveraging their generative diversity. The annotation team consisted of seven PhD candidates specializing in artificial intelligence and intelligent maritime navigation. The workflow proceeds through three strictly controlled stages:

\textbf{Stage 1. Human-Machine Collaborative Annotation:}  Annotators utilize a custom-developed GUI to process video clips collected from diverse hydrological environments, including rivers, lakes, canals, moats, harbors, and open seas. To ensure comprehensive cognitive coverage, we establish five hierarchical Q\&A categories: (1) Perception, (2) Scene Understanding, (3) Action/Interaction, (4) Causal \& Predictive, and (5) Knowledge-driven. The annotation followed a dual-path approach:
\begin{itemize}
    \item Manual Authoring: For complex scenarios requiring nuanced maritime logic (e.g., COLREGs interpretation), annotators directly authored Q\&A pairs.
    \item AI-Assisted Generation: To enhance efficiency and lexical diversity, annotators invoke a ChatGPT-5 agent via pre-defined system prompts targeting specific Q\&A categories. Crucially, acknowledging that general domain LLMs often suffer from hallucinations regarding professional waterway physics and rules, the annotators are required to strictly fact-check all AI-generated content against visual evidence. Only objectively accurate samples are retained for the next stage.
\end{itemize}

\textbf{Stage 2. Automated Grammatical Refinement \& Confirmation:} To eliminate linguistic ambiguities and grammatical errors without altering ground truth, the curated Q\&A pairs are processed by a secondary LLM agent dedicated to syntax refinement. Following this automated polish, the original annotators perform a mandatory confirmation review to ensure that the refined expressions remain faithful to the original maritime context and visual observations.

\textbf{Stage 3. Expert Validation \& Closed-Loop Feedback:} To guarantee the dataset's authority as a standard benchmark, we implement a multi-tier quality assurance mechanism. The final annotated pool is subjected to a blind audit by two external experts in intelligent navigation, distinct from the initial annotation team. This stage enforces a rigorous feedback loop: any sample flagged for factual inaccuracy, ambiguity, or subjectivity would be rejected and returned to the specific annotator for revision. A sample is only admitted into the final repository after achieving consensus between the annotator and the expert auditors.

\begin{figure*}
    \centering
    \centerline{\includegraphics[width=0.99\linewidth]{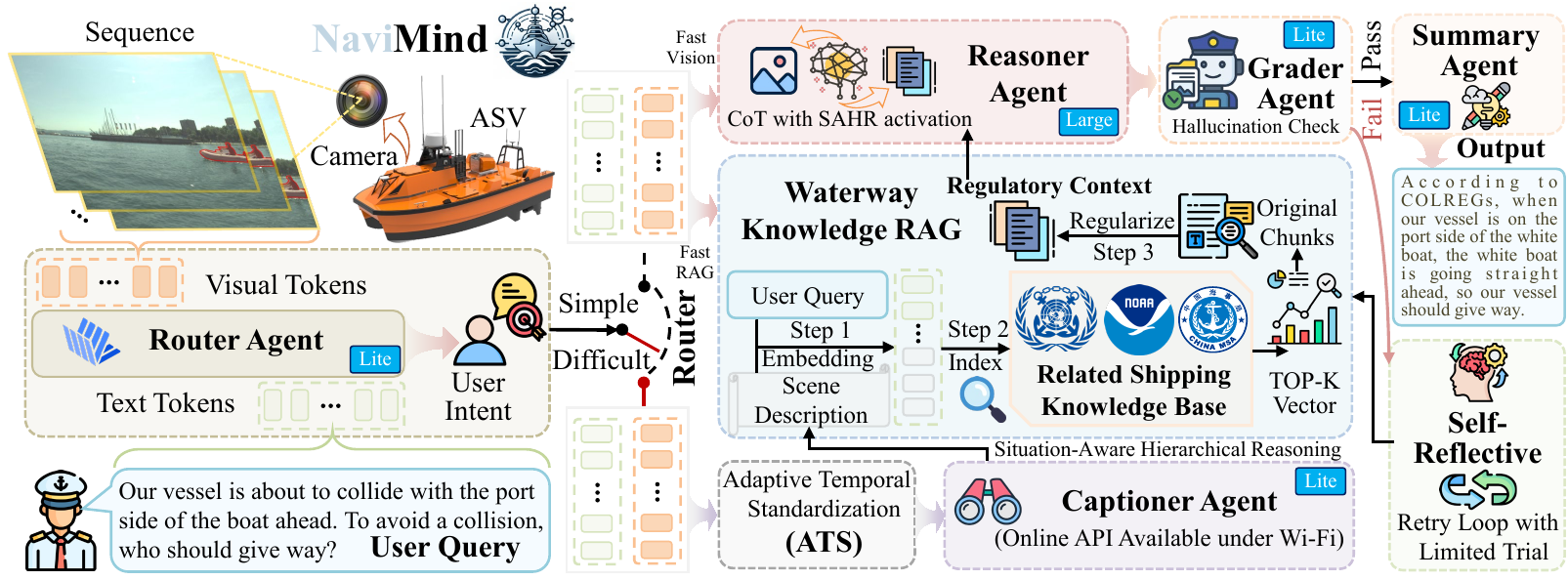}}
    \vspace{-3mm}
    \caption{The architecture of NaviMind, a multi-agent reasoning system for waterway navigation. NaviMind has two inputs (a user query and a video clip) and an output answer. It has 5 agents for routing, captioning, reasoning, grading and summary.}
    \label{fig:navimind_model}
\end{figure*}

Overall, to quantitatively validate the objectivity and consistency of our human-machine collaborative annotation pipeline, we conduct a rigorous Inter-Annotator Agreement (IAA) analysis. We randomly sample $10\%$ of the curated video-QA pairs and assign them to three independent maritime experts who were blinded to the original ground truths. These experts evaluate the generated answers based on three distinct criteria: visual grounding accuracy, regulatory compliance (e.g., strict alignment with COLREGs), and logical coherence.

We utilize Fleiss' Kappa ($\kappa$) to measure the statistical reliability of agreement across multiple raters. The overall Fleiss' Kappa score for the WaterVideoQA benchmark achieved 0.86, indicating almost perfect agreement in the academic standard. Specifically, the agreement on foundational tasks (e.g., Perception-based and Scene Understanding queries) reach an exceptional $\kappa$ of 0.92, while the highly complex Knowledge-Driven and Causal Reasoning tasks maintain a robust $\kappa$ of 0.81. These quantitative metrics rigorously demonstrate that our multi-stage, closed-loop annotation paradigm effectively mitigates subjective human biases and generative AI hallucinations, ensuring that WaterVideoQA serves as a highly reliable and objective benchmark for maritime cognitive evaluation.

\subsection{Statistics}
Fig. \ref{fig:dataset_stat} illustrates the statistical properties of WaterVideoQA. The dataset features a diverse distribution of QA pairs across five distinct types. Notably, over 50\% of the samples are dedicated to Action/Interaction (1,048 samples) and Causal \& Predictive (845 samples) tasks, reflecting a rigorous evaluation of a model's temporal reasoning capabilities. We also incorporate Perception-based and Scene Understanding queries to ensure foundational visual coverage, alongside a unique Knowledge-based category that tests rule-compliant logic. Linguistic analysis reveals rich semantic content, with mean lengths of 12.04 and 14.54 words for questions and answers, respectively. With video clips ranging from 1.33s to 126.4s (mean: 7.54s), the data captures sufficient temporal context for dynamic analysis across varying time scales. As evidenced by the detailed answer type statistics and the vocabulary word cloud, WaterVideoQA offers a broad spectrum of semantic concepts tailored to complex maritime interactions.

\section{Method}
\label{sec:method}
We present NaviMind, a holistic multi-agent framework designed to elevate ASV intelligence from passive perception to regulation-compliant cognitive reasoning. As illustrated in Fig. \ref{fig:navimind_model}, the architecture mirrors the human cognitive loop (perception, reasoning, and reflection) through three synergistic mechanisms, orchestrating a collaborative suite of five specialized agents, engineered to address the inherent challenges of efficiency, domain adaptability, and maritime safety.

\textbf{(1) Adaptive Semantic Routing (ASR)} serves as the initial cognitive dispatcher to resolve the critical trade-off between inference latency and reasoning depth . Governed by a lightweight Router Agent, it dynamically allocates computational resources by directing queries into optimized Fast Vision, Fast RAG, or Complex Reasoning pathways based on semantic intent.

\textbf{(2) Situation-Aware Hierarchical Reasoning (SAHR)} forms the cognitive core for high-order navigational tasks . Within this engine, Adaptive Temporal Standardization (ATS) first normalizes variable-length continuous video streams into condensed visual tokens . Assisted by a Captioner Agent for global scene contextualization, the central Reasoner Agent subsequently synthesizes these visual observations with retrieval-augmented maritime knowledge, enabling the system to perform physics-grounded and rule-compliant deductions.

\textbf{(3) Autonomous Self-Reflective Verification} provides a closed-loop reflection mechanism to mitigate the hallucination risks inherent in generative models . Managed by a Grader Agent, this module enforces rigorous self-correction to ensure logical consistency . Finally, upon passing verification, a Summary Agent distills the comprehensive reasoning chain into concise, actionable navigational guidance, ensuring that the ultimate decision is not only strictly aligned with professional safety standards but also readily executable.

\subsection{Adaptive Semantic Routing}
In real-world maritime operations, ASVs face a diverse array of inquiries ranging from instantaneous perceptual checks to complex regulatory reasoning. Employing a monolithic, heavy-weight reasoning pipeline for every query incurs unnecessary computational overhead and latency, which is detrimental to safety-critical edge computing platforms. To this end, we introduce the ASR, a mechanism to optimize cognitive resource allocation through semantic intent recognition.

Formally, let $q$ denote the user query and $v$ represent the real-time video stream. We model the routing process as a semantic classification task governed by a lightweight Router Agent (parameterized by $\theta_{\text{router}}$), which projects linguistic features of $q$ onto a decision space $\mathcal{P} = \{p_{\text{vis}}, p_{\text{rag}}, p_{\text{reason}}\}$, representing three distinct inference pathways. The optimal path indicator $d \in \mathcal{P}$ is determined by maximizing the conditional probability of the intent category:
\begin{equation}
d = \underset{c \in \mathcal{P}}{\arg\max} P(c \mid q; \theta_{\text{router}}).
\end{equation}
Based on the decision variable $d$, the system executes a conditional mapping function $\Phi(q, v, d)$ to generate the response $a$, which decouples simple perception from complex reasoning through the following strategies:

\begin{itemize}
    \item \textbf{Fast Vision Path ($p_{\text{vis}}$):} Activated for instant perceptual inquiries (e.g., "Is there a boat ahead?"). NaviMind bypasses knowledge retrieval and directly engages the Reasoner Agent, minimizing latency for high-frequency monitoring. \\
    \item \textbf{Fast RAG Path ($p_{\text{rag}}$):} Triggered for explicit knowledge-driven questions (e.g., "What does a green buoy signify?"). The router directs the flow solely to the Waterway Knowledge RAG to query the vector database $\mathcal{K}$, anchoring the response in maritime regulations without redundant visual processing. \\
    \item \textbf{Complex Reasoning Path ($p_{\text{reason}}$):} Reserved for high-order causal or predictive tasks (e.g., "Predict the collision risk based on current trajectories"). This path activates the full SAHR engine, which synthesizes visual features $v$, retrieved domain knowledge $\mathcal{R}$, and scene captions $\mathcal{C}_{\text{cap}}$ to perform multi-step deductive reasoning.
\end{itemize}

The final execution flow can be summarized as:
\begin{equation}
a = \Phi(q, v, d) = \begin{cases}
\mathcal{M}_{\text{vis}}(v, q), \ \text{if}\ d = p_{\text{vis}}, \\
\mathcal{M}_{\text{rag}}(q, \mathcal{K}), \ \text{if} \ d = p_{\text{rag}}, \\
\mathcal{M}_{\text{sahr}}(v, q, \mathcal{K}, \mathcal{C}_{\text{cap}}),  \text{if} \ d = p_{\text{reason}},
\end{cases}
\end{equation}
where $\mathcal{M}_{\text{vis}}$, $\mathcal{M}_{\text{rag}}$, and $\mathcal{M}_{\text{sahr}}$ denote specialized modules for visual perception, knowledge retrieval, and hierarchical reasoning, respectively. By filtering out simple queries from the heavy reasoning chain, ASR notably enhances the system's throughput and real-time responsiveness.

\subsection{Domain-Specific Maritime Knowledge Retrieval}
\label{subsec:rag}

Generic MLLMs inherently lack the specialized legal and physical intuition required for maritime navigation (e.g., COLREGs). To bridge this domain gap, we design a multi-modal Retrieval-Augmented Generation (RAG) module that structurally anchors the reasoning process in professional navigational constraints. As illustrated in the workflow, this module operates through three sequential mechanisms:

\textbf{Knowledge Base Regularization:} Let $\mathcal{D}$ denote the raw corpus of professional maritime regulations. We first parse and regularize $\mathcal{D}$ into a structured knowledge base $\mathcal{K} = \{k_1, k_2, ..., k_M\}$, where $M$ is the total number of discrete rule chunks. A pre-trained text encoder $\mathcal{E}_{text}$ is then employed to map each chunk into a high-dimensional semantic vector space, yielding $e_{k_i} = \mathcal{E}_{text}(k_i) \in \mathbb{R}^d$.
    
\textbf{Situation-Aware Query Embedding:} Standard RAG frameworks rely solely on the linguistic user query $q$, which is insufficient for dynamic embodied tasks. To address this, we introduce a situation-aware embedding strategy that seamlessly fuses $q$ with the global scene description $\mathcal{C}_{cap}$ (generated by the Captioner Agent). The multi-modal query embedding is formulated as:$$e_{query} = \mathcal{E}_{text}(q \oplus \mathcal{C}_{cap}) \in \mathbb{R}^d,$$
    where $\oplus$ denotes string concatenation. This ensures that the subsequent retrieval is explicitly conditioned on the real-time visual context, effectively filtering out irrelevant rules.

\textbf{Adaptive Indexing and Retrieval:} The system performs a dense similarity search between the fused situational query $e_{query}$ and the rule embeddings $e_{k_i}$. The retrieved regulatory context $\mathcal{R}_{kn}$ is constructed by selecting the Top-$K$ rule chunks that maximize the cosine similarity:$$\mathcal{R}_{kn} = \mathop{\arg\max}_{\mathcal{K}' \subset \mathcal{K}, |\mathcal{K}'|=K} \sum_{k_i \in \mathcal{K}'} \frac{e_{query} \cdot e_{k_i}}{\|e_{query}\| \|e_{k_i}\|},$$

These retrieved domain-specific constraints $\mathcal{R}_{kn}$ are subsequently injected into the SAHR engine, serving as hard logical boundaries to guide rule-compliant deductive reasoning.

\subsection{Situation-Aware Hierarchical Reasoning}
\begin{figure}[H]
    \centering
    \includegraphics[width=0.99\linewidth]{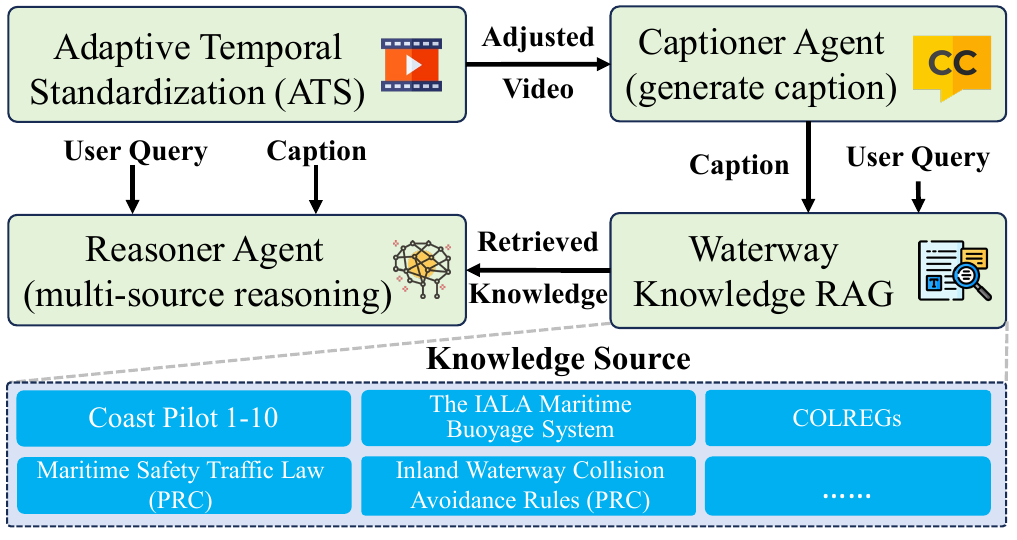}
    \vspace{-6mm}
    \caption{The workflow of Situation-Aware Hierarchical Reasoning. }
    \label{fig:sahr}
\end{figure}

While the routing architecture optimizes efficiency, the system's core cognitive capability for complex navigational tasks relies on the Situation-Aware Hierarchical Reasoning mechanism. Generic MLLMs inherently suffer from domain detachment, where they possess general visual recognition capabilities but lack the specific physical intuition and regulatory knowledge required for safety-critical maritime decision-making. SAHR addresses this by establishing a unified reasoning framework that mathematically enforces the alignment between visual perception and professional domain constraints. Formally, we model the reasoning process as a conditional generation task over an augmented semantic manifold.

Let the raw video stream captured by the ASV be denoted as $\mathcal{V} = \{f_i\}_{i=1}^{N}$, where $f_i$ is the $i$-th frame and $N$ represents the highly variable total number of frames across different maritime clips. To address this temporal heterogeneity and bound the computational complexity, we apply Adaptive Temporal Standardization (ATS) to dynamically extract a normalized set of $K$ key frames. The discrete sampling index set $\mathcal{I}$ is determined via a dynamic temporal projection function to ensure uniform motion coverage:
\begin{equation}
    \mathcal{I} = \left\{ \left\lfloor \frac{k-1}{K-1}(N-1) \right\rfloor + 1 \right\}_{k=1}^K.
\end{equation}
Subsequently, a visual encoder $\mathcal{E}_{vis}$ is employed to project the sampled key frames into a continuous embedding space, yielding the normalized visual token sequence $v$:
\begin{equation}
    v = \mathcal{E}_{vis}(\{f_j \mid j \in \mathcal{I}\}) \in \mathbb{R}^{K \times L \times D},
\end{equation}
where $L$ denotes the number of spatial patches flattened per frame, and $D$ is the feature embedding dimension. By transforming the raw video into this deterministic spatio-temporal representation $v$, we establish a standardized input for the reasoning engine.

Moreover, unlike standard VQA paradigms that directly model the posterior probability $P(A \mid v, q)$ given the visual sequence $v$ and the specific navigational query $q$, SAHR first constructs an explicit situation-aware context $\mathcal{S}$ to ground the inference. This involves two critical pre-processing steps:
\begin{enumerate}
\item \textbf{Global Captioning:} A descriptive context $\mathcal{C}_{\text{cap}}$ is generated to bridge the modality gap between pixel-level features and high-level semantics.
\item \textbf{Rule-Constraint Injection:} Relevant maritime regulations $\mathcal{R}_{\text{kn}}$ are retrieved from the vector knowledge base $\mathcal{K}$, explicitly conditioned on both the query and the scene description: $\mathcal{R}_{\text{kn}} = \text{Retrieve}(q, \mathcal{C}_{\text{cap}}, \mathcal{K})$ (Section \ref{subsec:rag}).
\end{enumerate}
The fused semantic context is thus defined as the concatenation of these heterogeneous modalities: $\mathcal{S} = v \oplus q \oplus \mathcal{C}_{\text{cap}} \oplus \mathcal{R}_{\text{kn}}$. The Reasoner Agent, parameterized by $\theta_{\text{reason}}$, generates the response sequence $A = \{y_1, y_2, \dots, y_T\}$ auto-regressively. The generation probability is formulated as:
\begin{align}
& P(A \mid \mathcal{S}; \theta_{\text{reason}}) \nonumber \\ 
& = \prod_{t=1}^{T} P(y_t \mid y_{<t}, \underbrace{v}_{\text{Visual}}, \underbrace{q}_{\text{Intent}}, \underbrace{\mathcal{C}_{\text{cap}}}_{\text{Scene}}, \underbrace{\mathcal{R}_{\text{kn}}}_{\text{Rules}}; \theta_{\text{reason}}),\label{eq:sahr_generation}
\end{align}
which enables a hierarchical reasoning capability that operates on two levels:
\begin{itemize}
\item \textbf{Level 1: Perceptual Grounding.} The model aligns visual tokens $v$ with regulatory definitions in $\mathcal{R}_{\text{kn}}$ (e.g., identifying a ``conical green buoy" not merely as an object, but as a ``Starboard Hand Mark").
\item \textbf{Level 2: Causal \& Predictive Deduction.} For high-order tasks, the model utilizes the full context $\mathcal{S}$ to perform Chain-of-Thought (CoT) reasoning. It extrapolates temporal dynamics from $v$ to predict future states (e.g., collision risks) that are strictly consistent with the navigational rules defined in $\mathcal{R}_{\text{kn}}$.
\end{itemize}
By explicitly conditioning the generative distribution on $\mathcal{R}_{\text{kn}}$, SAHR effectively minimizes hallucinations, ensuring that the ASV's decision-making is not only visually accurate but also interpretable and legally compliant.

\begin{algorithm}
   \caption{Self-Reflective Domain RAG Verification}
   \label{alg:sr_rag}
\begin{algorithmic}
   \STATE {\bfseries Input:} User Query $q$, Video $v$, Knowledge Base $\mathcal{K}$, Threshold $\tau$, Max Retries $N$
   \STATE {\bfseries Output:} Verified Answer $a^*$
   
   \STATE Initialize context $\mathcal{C} \leftarrow \text{Retrieve}(q, \mathcal{K})$
   \STATE Generate initial answer $a \leftarrow \text{Reasoner}(v, q, \mathcal{C})$
   \STATE Initialize retry counter $n = 0$
   \STATE Initialize $isVerified = false$
   
   \WHILE{$n < N$ {\bfseries and} $isVerified$ is $false$}
      \STATE Score $S \leftarrow \text{Grader}(q, a, \mathcal{C})$
      
      \IF{$S \ge \tau$}
         \STATE $isVerified = true$
      \ELSE
         \STATE $\mathcal{C}_{new} \leftarrow \text{ExpandRetrieval}(q, \mathcal{K})$
         \STATE Update context $\mathcal{C} \leftarrow \mathcal{C} \cup \mathcal{C}_{new}$
         \STATE Regenerate $a \leftarrow \text{Reasoner}(v, q, \mathcal{C})$
         \STATE $n \leftarrow n + 1$
      \ENDIF
   \ENDWHILE
   
   \STATE {\bfseries Return} $a$
\end{algorithmic}
\end{algorithm}

\begin{table*}
  \caption{Overall performances on the WaterVideoQA dataset. "nf" denotes no-finetuned while "f" denotes finetuned. The \textbf{Params} of Navimind indicates "(parameters of other agents + Reasoner Agent)".}
    \vspace{-3mm}
  \label{sample-table}
      \setlength\tabcolsep{3.1pt}
        \begin{tabular}{lc|ccccccccccc}
          \toprule
          \textbf{Methods}  & \textbf{Params}  & \textbf{ROUGE-1} & \textbf{ROUGE-2} & \textbf{ROUGE-L} & \textbf{BLEU-1} & \textbf{BLEU-2} & \textbf{BLEU-3} & \textbf{BLEU-4} & \textbf{METEOR} & \textbf{CIDEr} & \textbf{SPICE} & \textbf{GPT-Score}  \\
          \midrule
          InternVL3 (f) & 1B & 0.308 & 0.111 & 0.259 & 0.265 & 0.143 & 0.083 & 0.051 & 0.281 & 0.612 & 0.079 & 0.288 \\
          InternVL3 (f) & 2B & \textbf{0.316} & \textbf{0.113} & \textbf{0.262} & \textbf{0.269} & \textbf{0.170} & \textbf{0.082} & \textbf{0.043} & \textbf{0.303} & \textbf{0.328} & \textbf{0.098} & \textbf{0.307} \\
          \midrule
          InternVL3 (nf)  & 8B & 0.178 & 0.042 & 0.154 & 0.092 & 0.046 & 0.029 & 0.020 & 0.114 & 0.201 & 0.081 & 0.424 \\
          InternVideo 2.5 (f) & 7B & 0.351 & 0.125 & 0.274 & 0.273 & 0.179 & 0.090 & 0.074 & 0.308 & 0.369 & 0.139 & 0.443  \\
          QwenVL-3 (f) & 8B & 0.340 & 0.120 & 0.270 & 0.287 & 0.174 &0.102 & 0.090 & 0.310 & 0.388 & 0.151 & 0.450 \\
          MiniCPM-V 4.5 (f) & 8B & 0.349 & 0.127 & 0.279 & 0.289 & 0.175 & 0.107 & 0.086 & 0.319 & 0.399 & 0.150 & 0.459  \\
          \textbf{NaviMind (nf)} & (7+4)B & 0.331 & 0.121 & 0.267 & 0.281 & 0.172 & 0.083 & 0.060 & 0.311 & 0.384 & 0.152 & 0.452 \\
          \textbf{NaviMind (f)} & (7+4)B & \textbf{0.351} & \textbf{0.139} & \textbf{0.287} & \textbf{0.297} & \textbf{0.181} & \textbf{0.108} & \textbf{0.095} & \textbf{0.328} & \textbf{0.415} & \textbf{0.161} & \textbf{0.466} \\
          \midrule
          InternVL3 (nf)  & 14B & 0.255 & 0.088 & 0.217 & 0.185 & 0.104 & 0.067 & 0.046 & 0.216 & 0.321 & 0.136 & 0.456 \\
          OmAgent (nf) & 17B & 0.317 & 0.110 & 0.256 & 0.262 & 0.140 & 0.081 & 0.050 & 0.313 & 0.825 & 0.160 & 0.445 \\
          InternVL3 (f) & 14B & 0.368 & 0.150 & 0.290 & 0.301 & 0.191 & 0.134 & 0.101 & 0.335 &	0.627 & 0.189 & 0.479 \\
          \textbf{NaviMind (nf)} & (7+8)B &  0.359 & 0.162 & 0.303 & 0.319 & 0.190 & 0.128 & 0.110 & 0.330 & 0.851 & 0.205 & 0.483 \\
          \textbf{NaviMind (f)} & (7+8)B & \textbf{0.425} & \textbf{0.202} & \textbf{0.340} & \textbf{0.347} & \textbf{0.228} & \textbf{0.174} & \textbf{0.133} & \textbf{0.358} & \textbf{0.933} & \textbf{0.257} & \textbf{0.563} \\
          \midrule  
          VideoAgent (nf) & 24B & 0.329 & 0.137 & 0.271 & 0.288 & 0.181 & 0.118 & 0.102 & 0.318 & 0.841 & 0.191 & 0.471 \\ 
          InternVL3 (nf)  & 38B & 0.234 & 0.081 & 0.201 & 0.152 & 0.088 & 0.059 & 0.041 & 0.186 & 0.307 & 0.131 & 0.497 \\
          InternVL3 (f) & 38B & 0.392 & 0.178 & 0.327 & 0.315 & 0.203 & 0.125 & 0.118 & 0.337 & 0.795 & 0.231 & 0.524 \\     
          \textbf{NaviMind (nf)} & (7+14)B & 0.390 & 0.184 & 0.338 & 0.324 & 0.210 & 0.149 & 0.112 & 0.345 & 0.904 & 0.252 & 0.535 \\      
          \textbf{NaviMind (f)} & (7+14)B & \textbf{0.440} & \textbf{0.227} & \textbf{0.372} & \textbf{0.380} & \textbf{0.246} & \textbf{0.180} & \textbf{0.142} & \textbf{0.377} & \textbf{0.948} & \textbf{0.270} & \textbf{0.602} \\
          \bottomrule
        \end{tabular}
\end{table*}

\subsection{Autonomous Self-Reflective Verification}
Generative MLLMs inherently pose safety risks in maritime environments, where hallucinated maneuvers can cause severe collisions. To guarantee regulatory compliance, we introduce the Autonomous Self-Reflective Verification mechanism, a closed-loop feedback system driven by the Self-Reflective Domain RAG (SR-RAG) protocol. Specifically, a Grader Agent evaluates the Reasoner Agent's initial response against retrieved regulations, predicting a confidence score $S \in [0, 1]$ based on factual and logical consistency. If $S$ falls below a safety threshold $\tau$, an iterative feedback loop is triggered: the system expands the semantic retrieval scope for supplementary evidence and conditions the Reasoner Agent to regenerate its response. This paradigm ensures the final output is strictly aligned with professional standards rather than merely plausible, as detailed in Algorithm 1.

\section{Experiments}
\label{sec:experiments}

\subsection{Dataset Settings.} 
We first employ our proposed dataset WaterVideoQA for the baseline experiments, including 3,029 videos with 364.4K frames, accompanied with 3,673 QA pairs. WaterVideoQA has 2,573 training samples and 1,100 test samples. To evaluate the generalization of our proposed NaviMind, we also include LingoQA, another VideoQA datasets for road-based autonomous driving with 28K unique short video scenarios, and 419K annotations.

\subsection{Model Settings.} For WaterVideoQA, we evaluate models from two distinct paradigms: (1) End-to-end MLLMs, including Qwen3-VL, InternVL-3, InternVideo 2.5, and MiniCPM-V 4.5; (2) MAS, represented by VideoAgent and OmAgent. For our proposed NaviMind, we adopt a hierarchical configuration: the Router Agent utilizes lightweight 1B MLLM while the Grader, Captioner and Summary Agents employ 2B MLLMs, where the Captioner Agent employs local MLLM to ensure offline capability in connectivity-limited waterway scenarios. Conversely, the Reasoner Agent employs larger-scale models (4B, 8B, or 14B). InternVL3 serves as the default foundational MLLM of NaviMind. In VideoAgent and OmAgent, we also replace GPT-4 with InternVL3-14B.

\subsection{Training and Evaluation.} 
\textbf{Training.} For WaterVideoQA, we conduct full-parameter fine-tuning across all models using 8 A100 GPUs. During this process, the visual backbone remains frozen, while the LLM and projector layers are unfrozen and trainable. We utilize a global batch size of 8 (distributed as 1 sample per GPU) and resize input frames to a resolution of $448 \times 448$. The optimization is performed using AdamW with a weight decay of 0.01 and a warmup ratio of 0.03. The learning rate is initialized at $4 \times 10^{-5}$ and decayed via a cosine schedule. For LingoQA, we strictly adhere to the original hyperparameter settings, employing a learning rate of $5 \times 10^{-5}$, a total batch size of 8, and a weight decay of 0.1.

\textbf{Evaluation.} To comprehensively evaluate generation quality in terms of both sentence structure and semantic similarity, we employ a standard suite of text generation metrics. These include ROUGE-1/2/L, BLEU-1/2/3/4, METEOR, CIDEr, and SPICE. Besides, we incorporate GPT-Score to provide a robust assessment of semantic alignment and logical consistency.

\subsection{Quantitative Results}
\textbf{Overall Performances.} Table \ref{sample-table} details the quantitative results. NaviMind consistently outperforms competing methods across all metrics and model sizes. In the 11B parameter tier, NaviMind (f) exceeds leading MLLMs like MiniCPM-V 4.5 and InternVideo 2.5, securing the highest GPT-Score (0.466). When compared to specialized video agents, NaviMind (7+8)B significantly surpasses both OmAgent (17B) and VideoAgent (24B) in semantic accuracy, achieving a CIDEr score of 0.933 compared to their 0.825 and 0.841, respectively. Furthermore, scaling the Reasoner Agent to 14B yields the best overall performance (GPT-Score 0.602), validating the scalability of our architecture. Notably, even the non-finetuned NaviMind variants frequently outperform fully finetuned baselines, underscoring the intrinsic strength of the Situation-Aware Hierarchical Reasoning mechanism.

\begin{table}
    \centering
    \caption{Comparison of GPT-Score upon WaterVideoQA characteristics. \textbf{P}: perception; \textbf{S}: scene understanding; \textbf{C}: causal prediction; \textbf{A}: : action/interaction; \textbf{R}: knowledge-based reasoning.}
    \vspace{-3mm}
    \setlength\tabcolsep{4.5pt}
    \begin{tabular}{l|cccccc}
            \toprule
            \textbf{Methods}  & \textbf{P} & \textbf{S} & \textbf{C} & \textbf{A} & \textbf{R} & \textbf{Time(s)} \\
            \midrule
            OmAgent-17B & 0.460 & 0.388 & 0.451 & 0.352 & 0.438 & 19.78 \\
            VideoAgent-24B & 0.499 & 0.421 & 0.468 & 0.360 & 0.465 & 24.55 \\
            \midrule
            \textbf{NaviMind-11B} & 0.468 & 0.416 & 0.447 & 0.362 & 0.454 & \textbf{9.74} \\
            \textbf{NaviMind-15B} & 0.501 & 0.447 & 0.489 & 0.398 & 0.477 & 14.27 \\
            \textbf{NaviMind-21B} & \textbf{0.549} & \textbf{0.481} & \textbf{0.534} & \textbf{0.434} & \textbf{0.531} & 19.54 \\
            \bottomrule
        \end{tabular}
    \label{tab:ablation}
\end{table}

\begin{table}
    \centering
    \caption{Performances of various waterways.}
    \vspace{-3mm}
    \setlength\tabcolsep{5.0pt}
    \begin{tabular}{l|cccccc}
            \toprule
            \textbf{Methods} & \textbf{River} & \textbf{Lake} & \textbf{Canal} & \textbf{Moat} & \textbf{Harbor} & \textbf{Sea} \\
            \midrule
            NaviMind-11B & 0.452 & 0.475 & 0.415 & 0.422 & 0.458 & 0.490 \\
            NaviMind-15B & 0.485 & 0.508 & 0.445 & 0.455 & 0.492 & 0.512 \\
            NaviMind-21B & 0.538 & 0.565 & 0.495 & 0.505 & 0.542 & 0.568 \\
            \bottomrule
        \end{tabular}
    \label{tab:results_of_chara}
\end{table}

\begin{figure*}
    \centering
    \centerline{\includegraphics[width=0.98\linewidth]{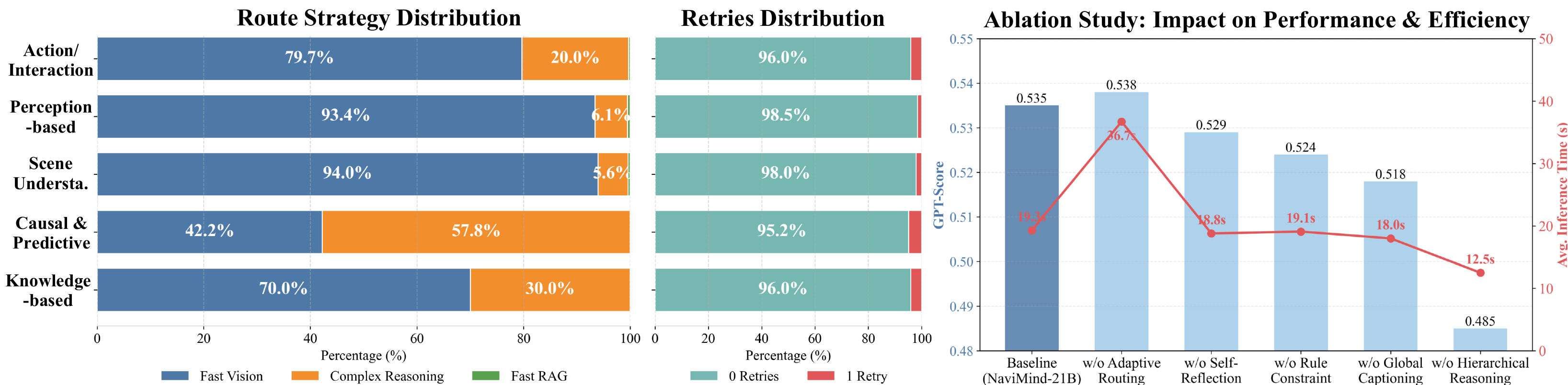}}
    \vspace{-3mm}
    \caption{Performance visualization of settings in NaviMind.}
    \label{fig:experiment_stats}
\end{figure*}

\textbf{Performance on Fine-grained Characteristics.} Table \ref{tab:ablation} validates NaviMind's superiority in both reasoning depth and efficiency. In cognitive tasks, NaviMind-21B establishes a new SOTA, notably outperforming the larger VideoAgent-24B in complex \textit{Causal} (0.534 vs. 0.468) and \textit{Action} domains. Crucially, NaviMind-11B offers exceptional efficiency with a 9.74s inference time, over 2$\times$ faster than competing agents, proving ideal for edge deployment. Environmentally (Table \ref{tab:results_of_chara}), the model exhibits robust generalization, maintaining high accuracy across both open waters (\textit{Sea}: 0.568) and constrained channels (\textit{Moat}: 0.505), effectively mitigating maritime domain gaps.

\textbf{Ablation Study and Mechanism Analysis.} We conduct a comprehensive analysis to validate the contribution of each NaviMind component, focusing on the trade-off between cognitive performance (GPT-Score) and inference latency.

\textbf{Impact of Architecture Components (Right):} The ablation results demonstrate that SAHR is the cornerstone of NaviMind's intelligence. Removing it causes the most precipitous drop in GPT-Score, reducing the system to a superficial perceptual model. Moreover, removing the ASR causes a massive latency surge with negligible performance gain, proving that our routing mechanism effectively bypasses redundant computation without compromising accuracy.

\textbf{Efficacy of Adaptive Semantic Routing (Left):} The usage distribution confirms the router's semantic intelligence. As shown in the ``Route Strategy Distribution", the agent correctly identifies task complexity: it dispatches over  of Perception and Scene Understanding queries to the lightweight Fast Vision path. Conversely, for high-order Causal \& Predictive tasks, it dynamically shifts load, directing  of queries to the Complex Reasoning path (Orange). This selective activation ensures optimal resource utilization.

\textbf{Efficiency of Self-Reflective Verification (Center):} The ``Retries Distribution" highlights the high ``first-pass" accuracy of our Reasoner Agent. The system achieves a 0-Retry rate of across all categories, meaning the Verification mechanism acts as a precise ``safety net" rather than a bottleneck. Although the system allows for multiple iterations, empirical data shows that at most 1 retry is triggered to correct hallucinations, ensuring that safety checks do not impose a systemic latency penalty.

\begin{table}
  \centering
  \caption{Generalization experiments on the LingoQA dataset.}
    \vspace{-3mm}
  \label{sample-table}
      \setlength\tabcolsep{1.3pt}
        \begin{tabular}{l|c|c|c|cccc}
          \toprule
          \textbf{Methods}  & \textbf{Categories} & \textbf{No. Frames} & \textbf{Lingo-J} & \textbf{BLEU-4} & \textbf{METEOR} & \textbf{CIDEr} \\
          \midrule
            LingoQA & zero-  & 5 & 33.6 & 8.33 & 14.33 & 39.16 \\
            VideoAgent & shot & 5 & 54.18 & 11.79 & 18.77 & 42.64\\
            \textbf{NaviMind} & models & 5  & \textbf{60.42} & \textbf{13.76} & \textbf{21.88} & \textbf{46.63} \\       
          \midrule
            LingoQA & fine-  & 5 & 60.80 & 15.00 & 18.56 & 65.62 \\
            VideoAgent & tuned & 5 & 68.39 & 20.07 & 26.65 & 85.92\\
            \textbf{NaviMind} & models & 5 & \textbf{72.55} & \textbf{23.67} & \textbf{31.41} & \textbf{93.77} \\
          \bottomrule
        \end{tabular}
\end{table}

\textbf{Generalization on LingoQA.} To validate the architectural universality of NaviMind beyond maritime scenarios, we evaluate it on the autonomous driving benchmark LingoQA. \textbf{(1) Zero-Shot Robustness:} NaviMind exhibits exceptional intrinsic adaptability, achieving a Lingo-J score of 60.42 without domain-specific training, significantly outperforming both VideoAgent ($54.18$) and the LingoQA baseline ($33.6$). \textbf{(2) Transfer Learning Capability:} Upon fine-tuning, NaviMind establishes new state-of-the-art performance with a CIDEr of 93.77 and Lingo-J of 72.55, surpassing the competitive VideoAgent by notable margins (+7.85 CIDEr). This confirms that our SAHR mechanism effectively captures universal navigational logic applicable to general embodied agents.

\begin{figure*}
  \begin{center}
    \centerline{\includegraphics[width=0.98\linewidth]{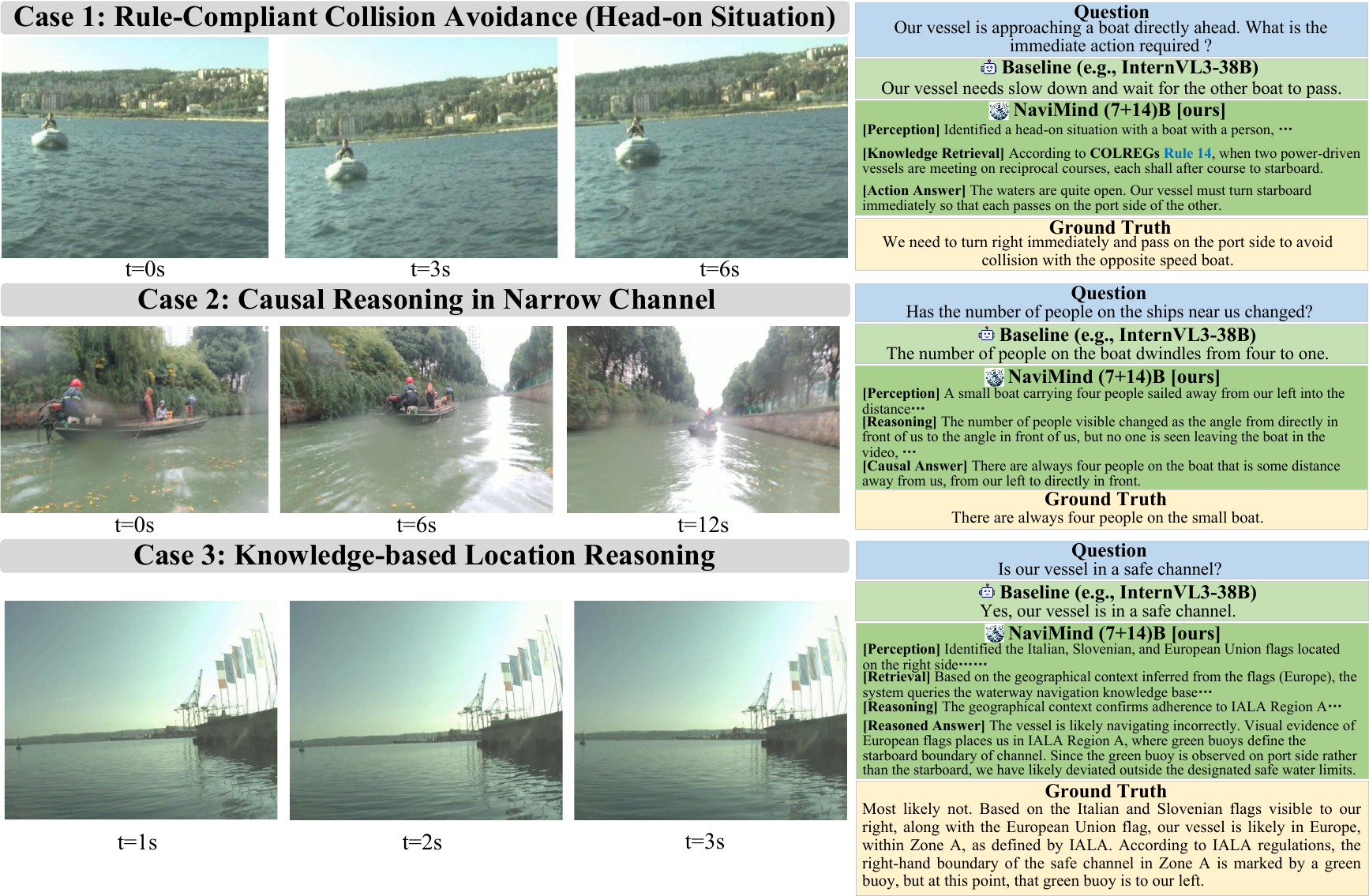}}
    \vspace{-3mm}
    \caption{Visualization of prediction by InternVL3 and our proposed NaviMind.}
    \label{fig:vis_pred}
  \end{center}
\end{figure*}

\textbf{Qualitative Analysis.} The visualization (Fig. \ref{fig:vis_pred}) highlights a critical disparity in cognitive depth between the systems: while InternVL-3 relies on superficial visual pattern matching, NaviMind demonstrates grounded professional reasoning. In the safety-critical collision avoidance scenario (Case 1), NaviMind explicitly anchors its decision in COLREGs Rule 14, generating a precise, regulation-compliant maneuver (``turn starboard") that contrasts sharply with the baseline's passive and vague suggestion. Furthermore, in the temporal consistency task (Case 2), our system employs causal deduction to maintain object permanence across perspective shifts, effectively correcting the counting hallucination exhibited by InternVL-3. This confirms NaviMind's ability to provide interpretable, legally grounded, and logically robust navigational guidance.

\begin{figure}
  \begin{center}
    \centerline{\includegraphics[width=0.98\linewidth]{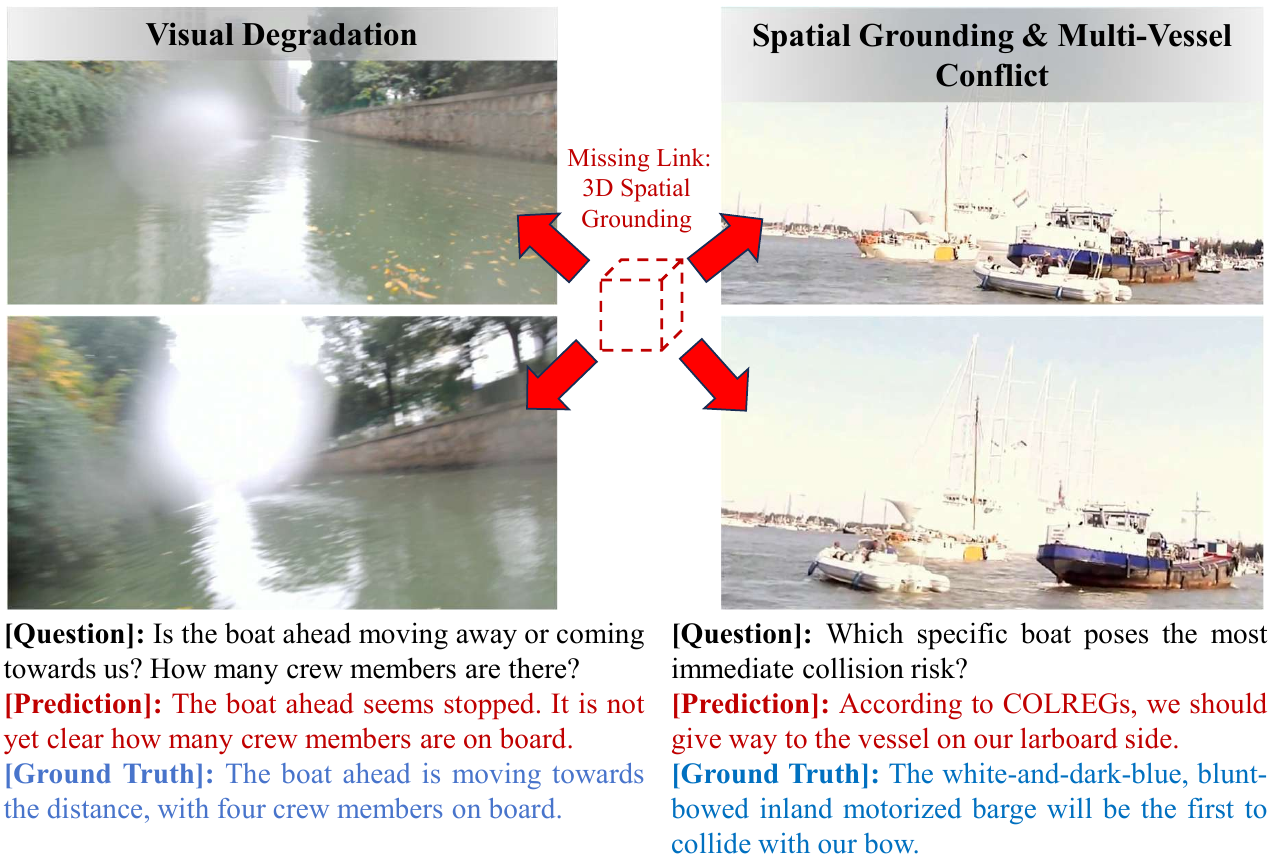}}
    \vspace{-3mm}
    \caption{Visualization of the limitations.}
    \label{fig:limitation}
  \end{center}
\end{figure}

\section{Limitations and Future Works}
\label{sec:limit}

Although NaviMind establishes a robust neuro-symbolic baseline for maritime cognitive reasoning, several constraints remain for deployment in fully autonomous, safety-critical operations. As Fig. \ref{fig:limitation} presents,
\textbf{first, }relying exclusively on visible-spectrum video exposes the framework to severe degradation under extreme maritime conditions (e.g., dense fog, nighttime glare). Such optical ambiguity can corrupt visual tokenization, risking cascading reasoning failures. Future work will evolve WaterVideoQA into a multi-sensor benchmark by fusing 4D millimeter-wave radar and LiDAR, enabling all-weather, geometry-aware perception.
\textbf{Second, }while NaviMind excels in rule-compliant semantic deduction, it lacks fine-grained spatial grounding. NaviMind formulates actions based on global context but cannot explicitly output dense trajectory coordinates or instance-level masks. To bridge this, we plan to integrate a specialized Spatial-Grounding Agent to map textual deductions directly onto pixel-level navigable space.
\textbf{Finally, }resolving multi-vessel game-theoretic interactions remains challenging. While the RAG effectively regularizes pairwise encounters via COLREGs, complex topological conflicts involving multiple vessels strain purely auto-regressive decoding. Future research will explore multi-agent reinforcement learning to simulate counterfactual trajectory planning and optimize long-horizon collision avoidance policies.

\section{Conclusion}
\label{sec:conclusion}

In this paper, we address the critical disparity between passive visual perception and active cognitive reasoning in autonomous maritime navigation. We introduce WaterVideoQA, the first comprehensive video question-answering benchmark tailored for all-waterway environments, which rigorously evaluates hierarchical cognitive capabilities ranging from basic perception to causal deduction. To tackle the challenges of computational efficiency and safety compliance on edge devices, we propose NaviMind, a novel multi-agent neuro-symbolic framework. By synergizing Adaptive Semantic Routing for optimized resource allocation and Situation-Aware Hierarchical Reasoning for regulation-grounded deduction, NaviMind effectively bridges the gap between visual observations and professional maritime rules. Extensive experiments demonstrate that NaviMind not only establishes new state-of-the-art performance on WaterVideoQA but also delivers superior inference efficiency and robust cross-domain generalization. This work paves the way for the development of trustworthy, interpretable, and logically robust cognitive systems for next-generation ASVs.


%




\ifCLASSOPTIONcaptionsoff
  \newpage
\fi



\footnotesize
\bibliographystyle{IEEEtran}
\bibliography{bare_jrnl}
\end{document}